\documentclass[letterpaper, 10 pt, conference]{ieeeconf}  

\IEEEoverridecommandlockouts

\usepackage{graphicx} 
\usepackage{amsmath} 
\usepackage{amssymb}  
\usepackage[dvipsnames]{xcolor}
\usepackage{xspace}
\usepackage{url}

\usepackage{array}
\usepackage{booktabs}

\usepackage{hyperref}
\usepackage{subfigure}
\usepackage{bytefield}
\usepackage{multirow}
\usepackage{tikz}
\usepackage{rotating}
\usetikzlibrary{positioning, shapes, arrows, automata}
\usepackage{glossaries}
\glsdisablehyper 
\usepackage[normalem]{ulem}
\usepackage{fancyhdr}

\usepackage{marginnote}

\fancypagestyle{withfooter}{
  
  \fancyfoot[C]{\footnotesize Accepted to the IEEE ICRA Workshop on Field Robotics 2024}
}

\hypersetup{
    colorlinks,
    linkcolor={red!50!black},
    citecolor={blue!50!black},
    urlcolor={blue!80!black}
}

\newif\ifdraftcolor

\ifdraftcolor
\newcommand{\fer}[1]{{\color{orange}#1}}

\newcommand{\todo}[1]{{\color{red}#1}}

\newcommand{\deletecamready}[1]{{\color{Red}\sout{#1}}}
\newcommand{\authorlist}{Authors}
\else
\newcommand{\fer}[1]{#1}

\newcommand{\todo}[1]{}

\newcommand{\deletecamready}[1]{}
\newcommand{\authorlist}{Fernando Cladera*, Ian D. Miller*, Zachary Ravichandran,  Varun Murali, Jason Hughes,\\ M. Ani Hsieh, C. J. Taylor, and Vijay Kumar}
\fi

\newcommand{\citationneeded}[1]{$^{[\text{\color{blue}citation needed}]}$~}

\newacronym{cots}{COTS}{commercial-off-the-shelf}
\newacronym{manet}{MANET}{mobile ad-hoc mesh network}
\newacronym{uav}{UAV}{unmanned aerial vehicle}
\newacronym{phy}{PHY}{physical layer}
\newacronym{snr}{SNR}{signal-to-noise ratio}
\newacronym{rtt}{RTT}{round-trip time}
\newacronym{rssi}{RSSI}{received signal strength indicator}
\newacronym{ugv}{UGV}{unmanned ground vehicle}
\newacronym{ntp}{NTP}{network time protocol}
\newacronym{asoom}{ASOOM}{Aerial Semantic Online Ortho-Mapping}
\newacronym{mocha}{MOCHA}{Multi-robot Opportunistic Communication for Heterogeneous Collaboration}
\newacronym{bvlos}{BVLOS}{Beyond Visual Line of Sight}

\title{\LARGE \bf
Challenges and Opportunities for Large-Scale Exploration\\with Air-Ground Teams using Semantics
}

\author{\authorlist%
\thanks{All authors are with GRASP Laboratory, University of Pennsylvania. 
*These authors contributed equally. Corresponding author: {\tt\small fclad@seas.upenn.edu}.}
\thanks{
We gratefully acknowledge the support of
ARL DCIST CRA W911NF-17-2-0181, 
NSF Grants CCR-2112665, 
ONR grant N00014-20-1-2822, 
ONR grant N00014-20-S-B001, 
NVIDIA, 
and C-BRIC, a Semiconductor Research Corporation Joint University Microelectronics Program program cosponsored by DARPA. 
}
}

\begin{document}

\maketitle
\thispagestyle{withfooter}
\pagestyle{withfooter}

\begin{abstract}
One common and desirable application of robots is exploring potentially hazardous and unstructured environments.  Air-ground collaboration offers a synergistic approach to addressing such exploration challenges.
In this paper, we demonstrate a system for large-scale exploration using a team of aerial and ground robots.
Our system uses semantics as \emph{lingua franca}, and relies on fully opportunistic communications.
We highlight the unique challenges from this approach, explain our system architecture and showcase lessons learned during our experiments.
All our code is open-source, encouraging researchers to use it and build upon\footnote{The code for the different components of this paper is accessible at \url{https://github.com/KumarRobotics/spomp-system}.}.
\end{abstract}


\glsresetall

\section{Introduction}
The exploration of large-scale, unknown and unstructured environments is a challenging task that can significantly benefit from the use of robots. Depending on a robot's mobility capabilities and the environment, different types of vehicles can be used such as \glspl{uav} and \glspl{ugv}. Each robot type offers distinct advantages and disadvantages, motivating the use of heterogeneous teams that can leverage their complementary strengths. Coordinated multi-robot teams can decrease the overall exploration time, increase mission capabilities, and extend exploration ranges compared to single-robot operations.

As the size of the region of interest grows, deploying multiple robots becomes increasingly challenging.  However, with an increasing number of robots, the probability of individual robot failures also rises, potentially compromising the mission's success. Given this trade-off, there is a need to develop strategies that can effectively perform large-scale exploration while being robust to individual failures. 

In this paper, we explore the challenges and opportunities of large-scale exploration using air-ground robot teams. We describe our approach to addressing various challenges, including robot coordination and communication, planning, and mapping. We present potential paths for improvement, and discuss lessons learned in implementing both the real-world and simulation experiments of our system. 


\begin{figure}
    \centering
    \includegraphics[width=.49\linewidth]{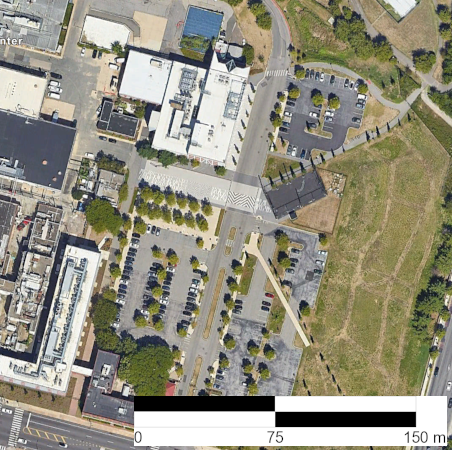}
    \includegraphics[width=.49\linewidth]{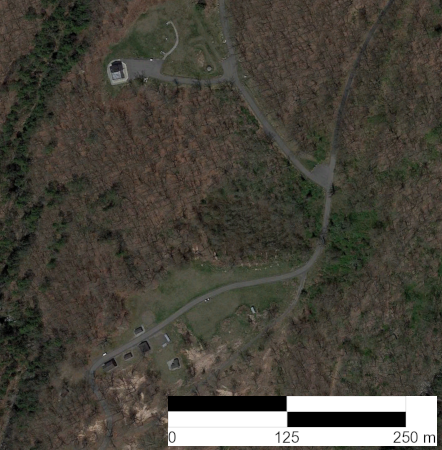}

    \vspace{.1cm}
    \includegraphics[width=.745\linewidth]{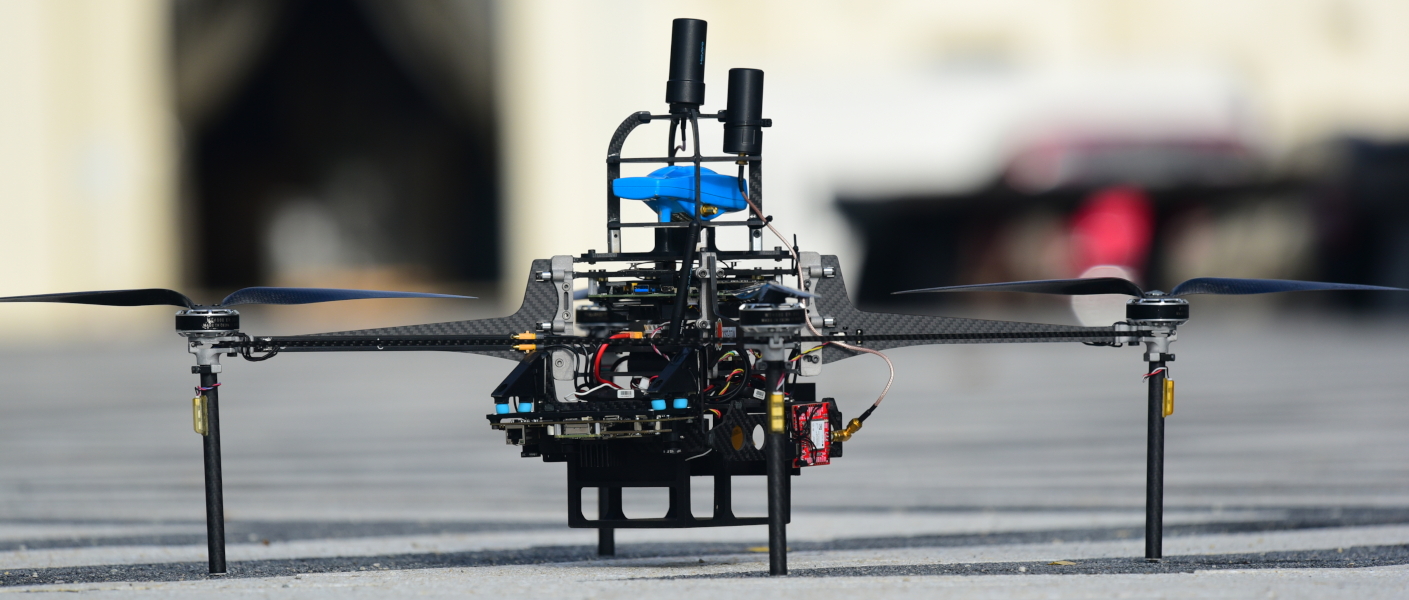}
  \includegraphics[width=.24\linewidth]{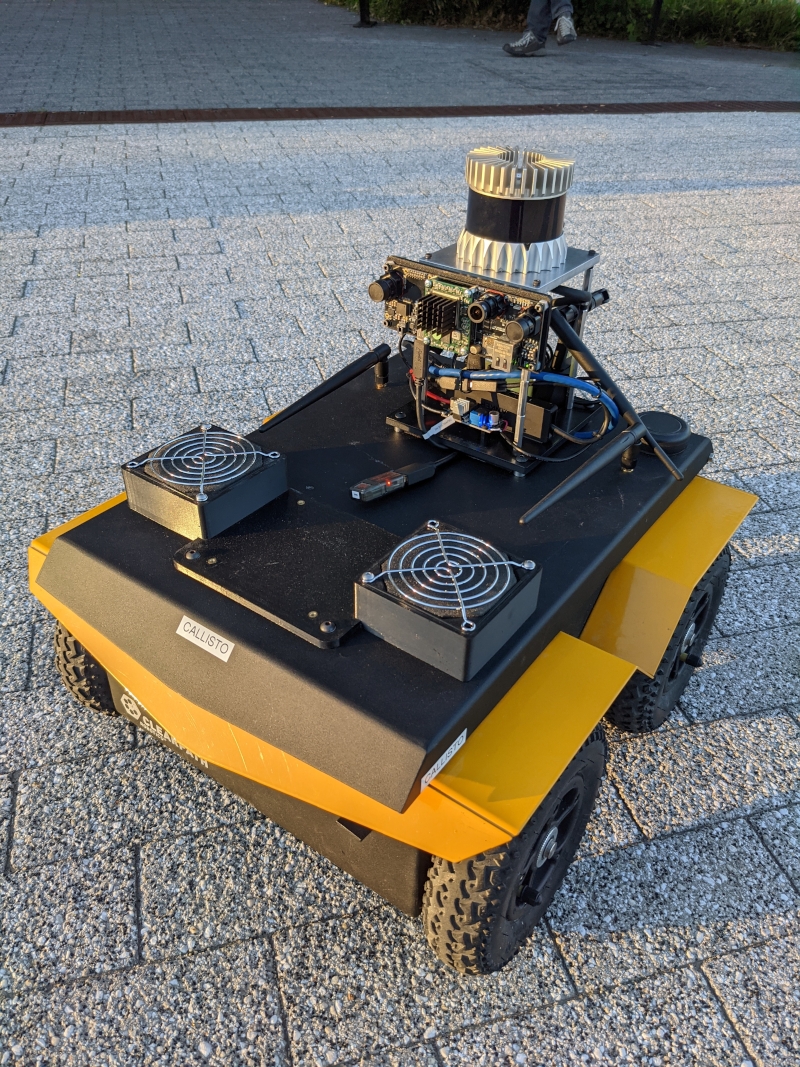}
    \caption{\uline{Top}: the environments where we performed exploration were as big as $157000\,m^2$ ($15.7\,\text{ha}$) including urban (left) and rural settings (right). \uline{Bottom Left}: high-altitude \gls{uav} used to perform online mapping and communications.
    \uline{Bottom Right}: Jackal robots using during the experiment. The LiDAR was used for navigation, state estimation, and obstacle avoidance. Cameras were used only for data logging purposes. Figures from~\cite{cladera2023enabling}.}
    \label{fig:fig1}
    \vspace{-.5cm}
\end{figure}
\section{Related Work}
The exploration of unknown areas using multiple robots is a longstanding approach in field robotics aimed at reducing exploration time~\cite{guzzoni1997many, burgard2005coordinated}, providing redundancy in case of robot failure~\cite{ebadi2024present}, or establishing active communication links between the operator and the robots~\cite{stump2011visibility}.

\glspl{uav} flying at high altitudes can provide a bird's-eye viewpoint, cover greater distances, and move at higher speeds due to the lack of obstacles. \Gls{cots} platforms are available for mapping, and orthomaps can be generated with open-source tools such as OpenDroneMap~\cite{OpenDroneMap}.
However, \glspl{uav}, particularly multicopters, are limited by their high energy requirements for hovering which translate into short operation times.
\Glspl{ugv}, on the other hand, are more energy efficient, can carry more sensors and heavier payloads, and can safely get near objects of interest. 
Their locomotion is limited by the terrain characteristics, and they therefore need to avoid obstacles and plan trajectories accordingly.  

It is common to take advantage of the benefits of both platforms for large scale exploration; \glspl{uav} can scout the terrain rapidly, looking for objects of interest. \glspl{ugv} can then be deployed for closer inspection or validation. These approaches have been followed in the literature by numerous authors, but they usually assume constant communication~\cite{peterson2018online}, explore limited areas~\cite{grocholsky2006cooperative}, or require offline processing to generate maps~\cite{gabrlik2021automated}.

Finally, communication plays an important role in air-ground exploration. Some works focus on ensuring continuous network connectivity, deploying robots to maintain communication among the different nodes of the network~\cite{hsieh2008maintaining, mox2022learning}. However, for large areas, all-time connectivity may not be feasible or too limiting. As such, it makes sense to enable robots to communicate more opportunistically, establishing intermittent links and exchanging information with other nodes when they are in communication range. Some approaches for opportunistic communication were developed for the DARPA SubT challenge~\cite{Ginting2021CHORDDD, saboia2022achord}, but they require periodic rendezvous between robots to exchange information. Other approaches ensure future rendezvous for successful communication~\cite{xi2020synthesis}.

It is additionally important to consider the form of information being communicated between heterogeneous agents.
\Glspl{uav} perceive the environment very different from \glspl{ugv}, and must reconcile these differences to extract actionable information.
We argue that \emph{semantics} are ideally suited for this role due to their invariant properties~\cite{bowman2017probabilistic}.
We therefore make semantics a key component of both the mapping algorithm running on the \gls{uav} as well as the  planning and obstacle avoidance components of the \glspl{ugv}.

In this work, we showcase the approaches we used for multi-robot exploration of large-scale environments.
We emphasize our use of semantics throughout, as well as our focus on the real-time operation of all components of the system.
Moreover, we will show how communications can be fully opportunistic, without requiring periodic rendezvous~\cite{xi2020synthesis} or ensured network connectivity~\cite{hsieh2008maintaining, mox2022learning}. 

\section{Methods}
\begin{figure*}
    \centering
    \includegraphics[width=.7\linewidth]{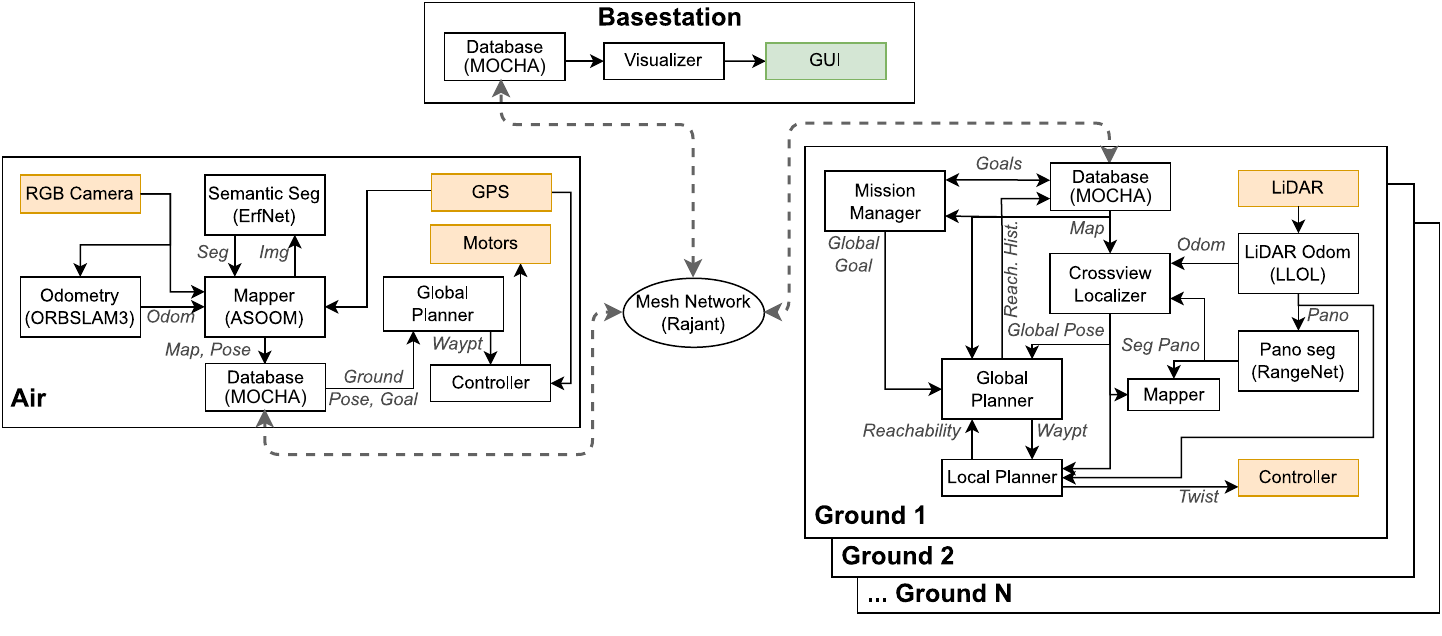}
    \caption{System architecture. Sensors are shown in orange and visualization outputs in green. Figure from~\cite{miller2024spomp}.}
    \label{fig:system}
    \vspace{-.5cm}
\end{figure*}

This section provides an overview of the systems architecture, shown in Fig.~\ref{fig:system}. These systems are described in detail in multiple previous works and we refer the interested reader to~\cite{cladera2023enabling, miller2022stronger, miller2024spomp}.  We provide a brief summary below. 

\subsection{Systems}

\subsubsection{Ground Robots}
We based our platforms on the Clearpath Jackal. 
The Jackals were retrofitted with an AMD Ryzen 3600 CPU and an NVIDIA GTX 1650 GPU. 
The sensor payload of the robots include an Ouster OS1-64 LiDAR used to perform navigation and obstacle avoidance, and a Realsense D435 or an Open Vision Computer  (OVC)~\cite{falcon250ovc} for data logging purposes. An example of this platform is shown in Fig.~\ref{fig:fig1}.

\subsubsection{Aerial Robots}
Our main platform is based on the Falcon 4 \gls{uav}~\cite{miller2022stronger}, fitted with an onboard OVC, a U-blox F9P GPS module, and a PX4 flight controller. 
The \glspl{uav} carries an Intel NUC with 32 GB of RAM and an i7-10710U processor. The OVC does not perform any preprocessing and just provides global-shutter RGB images. The \gls{uav} is shown in Fig.~\ref{fig:fig1}.

All computations are performed onboard for both aerial and ground platforms.

\subsubsection{Communication Physical Layer}
We build our communication approaches on the Rajant \gls{manet}. However, we perform only point-to-point communications (one-hop) to reduce communication bandwidth. 
Aerial and ground robots are equipped with DX2 modules, and we use an ME4 module for the base station.

\subsection{Aerial Mapping}
\begin{figure}[b]
    \centering
    \includegraphics[width=\linewidth]{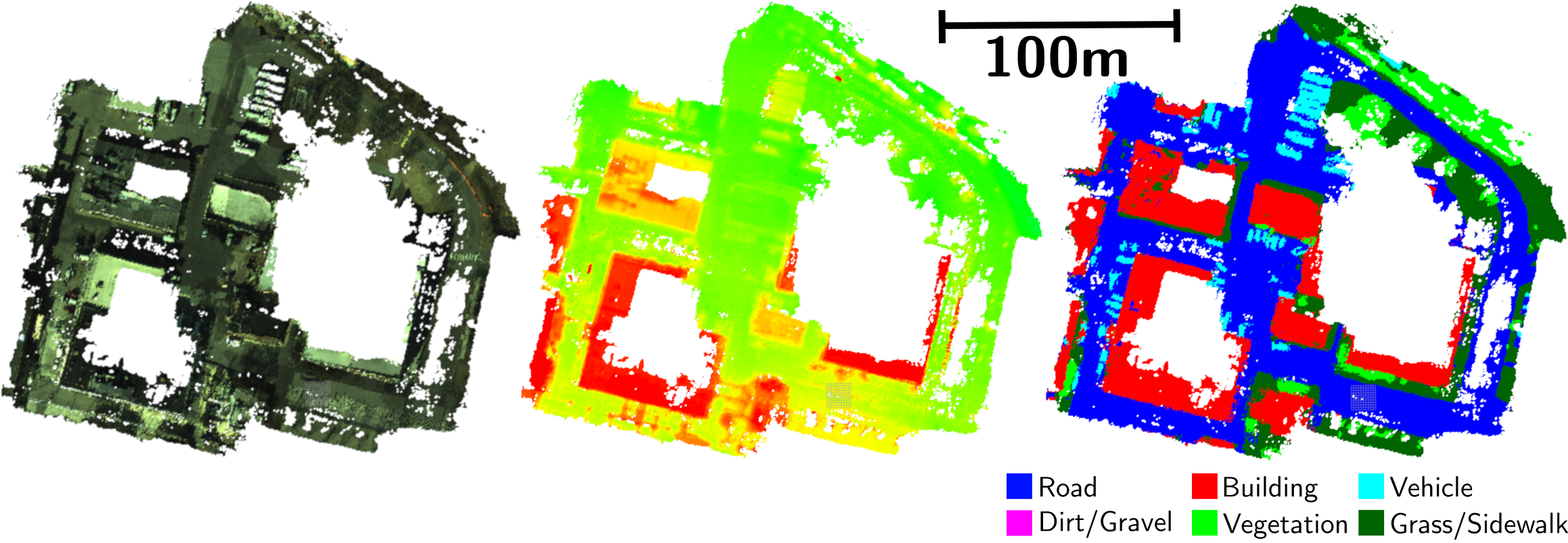}
    \caption{\gls{asoom} outputs computed onboard the \gls{uav}: orthomap, elevation map, and semantic map. Figure from~\cite{miller2022stronger}.}
    \label{fig:asoom}
    \vspace{-.5cm}
\end{figure}

The \gls{uav} performs online mapping using the \gls{asoom} module, described in~\cite{miller2022stronger}. This module relies on ORBSLAM3~\cite{ORBSLAM3_TRO} to obtain the poses of the UAV at regularly spaced keyframes.
It also uses ErfNet~\cite{romera2018erfnet} to segment the keyframe images.
\Gls{asoom} then constructs a point cloud map using block matching, which is colored and segmented using the RGB image and semantic segmentation. 
An example of the \gls{asoom} output can be seen in Fig.~\ref{fig:asoom}. The segmented classes correspond to building, roads, grass/sidewalk, vegetation, dirt/gravel and vehicles.

\subsection{Ground Robot Planning}

\Glspl{ugv} use LLOL~\cite{qu2022llol} to obtain odometry and integrate LiDAR scans into depth panoramas. 
These panoramas are segmented by RangeNet++~\cite{milioto2019rangenet} and used to cross localize the \glspl{ugv} in the \gls{uav} map by matching semantics.
We also perform terrain characterization in panorama space.
Robots analyze the depth panorama to compute local gradients, which inform traversability that is used for local planning.

The global planner of the \glspl{ugv} is based on the aerial map generated by the \gls{uav}. Ground robots infer traversability from the aerial map and their own local analysis. They also share their experiences with other ground robots. We refer the reader to our work \cite{miller2024spomp} for more details.

\subsection{Communications}
\Gls{mocha} was used for communications. \Gls{mocha} is a gossip-based protocol that provides fully opportunistic communication between the different nodes of the system. \gls{mocha} does not require periodic rendezvous, and enables robots to act as \emph{data mules}. This system is described in detail in~\cite{cladera2023enabling}.

\gls{mocha} also provides information and metrics about the communication links with different robots, such as the last synchronization time, link quality, and status of communications. These metrics can be used by the robots to inform high-level planning algorithms.

\subsection{UAV Planning}
The \gls{uav} performs a dual exploration/communication mission: it scouts the area and constructs a semantic map, but it also acts as a data mule, carrying messages between different ground robots. The transition between these two states is determined by predefined time allocations, or by the status of the communication reported by \gls{mocha}. We refer the reader to~\cite{cladera2023enabling} for more details.

\section{Experiments, Lessons Learned, and Opportunities}
\subsection{Simulation Experiments}

We performed simulations in a physics-based photorealistic simulator built on Unity, with the objective of evaluating the performance of the system as the number of agents increases~\cite{cladera2023enabling, miller2022stronger} and performing ablations of different components of our system.

We also modeled communication latency based on real-world data and included these models in the simulator.  This approach enables us to evaluate communication contention. Contention can be a deterministic factor when the number of agents is high and the communication radii is comparable to the environment size.

\textbf{Lessons Learned} 
\subsubsection{Number of Robots and Configuration} In both~\cite{cladera2023enabling} and~\cite{miller2022stronger} we observed that increasing the number of \glspl{ugv} improved performance, measured as the number of goals reached by the \glspl{ugv} and the average time to goal. 
When the \gls{uav} only performs exploration, increasing the number of \glspl{ugv} produces diminishing returns~\cite{miller2022stronger}. When using a \gls{uav} that performs both exploration/communication, the system scales well when the number of robots increase.

\subsubsection{Abstraction Level}
\fer{When simulating multiple agents, choosing the right level of abstraction can help scale simulations. 
In our simulation experiments we avoided some perception and state estimation tasks that are difficult to transfer to real-world experiments and require significant compute. 
For instance, instead of running segmentation and odometry in-the-loop, we directly simulated the outputs of these algorithms.
This allowed us to employ the simulator to debug the high level state machines and robot interactions, as well as test with larger teams.}

\textbf{Opportunities} 
\setcounter{subsubsection}{0}
\subsubsection{Data-driven policies} The policies used by the ground and aerial robots in both simulation and real-world experiments were manually tuned in the simulator. Nonetheless, data-driven approaches could be used to learn better policies or task allocations for the different robots. For instance, the \gls{uav} transitions between exploration and communication occur after a predefined timeout, which may not be optimal for a particular environment configuration.
\subsubsection{Intelligent Goal Assignment}
The robots used a greedy approach when selecting goals to visit.
This frequently led to suboptimal routing, where a robot would drive right over a goal to visit a different one, but because it was not technically \emph{targeting} the goal another robot would have to visit later.
More intelligent distributed coordination would have a significant effect on the efficiency of our system.

\subsection{Real-World Experiments}

We demonstrated our system with 1 \gls{uav} and 3 \glspl{ugv}, in different large-scale environments ranging from $62500\,\text{m}^2$ ($6.25\,\text{ha}$) to $157000\,\text{m}^2$ ($15.7\,\text{ha}$), in both urban and rural settings. Our \glspl{ugv} traveled a total of 17.8 km, with the robots spending most of the time in autonomous mode. 

\begin{figure}
    \centering
    \includegraphics[width=\linewidth]{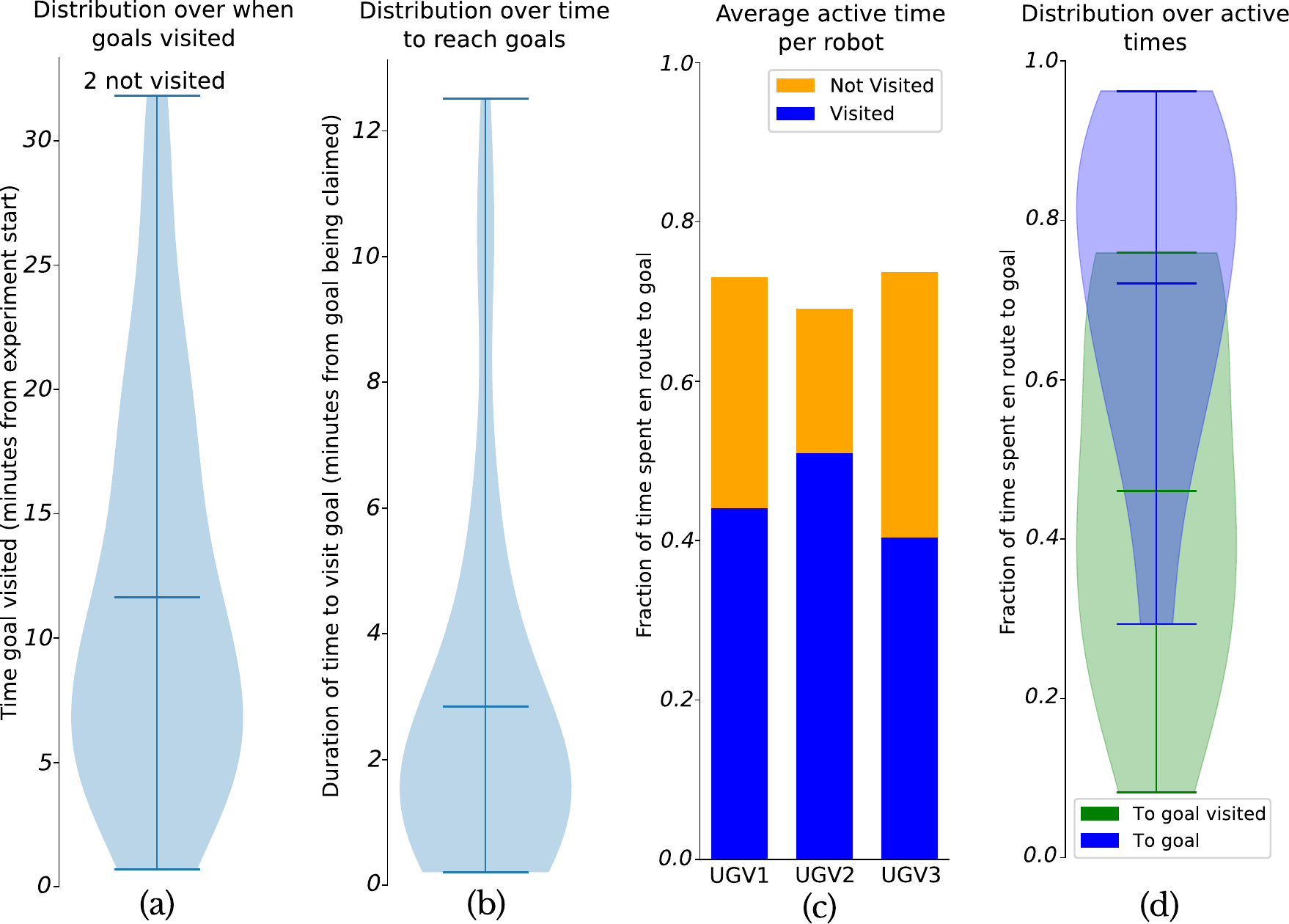}
    \caption{Analysis of real-world experiments from the \gls{ugv} point-of-view. Most goals were visited during the first half of the experiment, with some goals being difficult to visit. Only two goals were not visited.  Goals were roughly evenly distributed between robots, and robots spent more than 75\% of the time navigating to a goal, on average. Figure from~\cite{miller2024spomp}.}
    \label{fig:experiment-results}
    \vspace{-.5cm}
\end{figure}

In most experiments, the \glspl{ugv} visited all the goals obtained from the \gls{uav} map, with a total of 49 goals visited out of the 51 goals. A summary of these experiments is shown in Fig.~\ref{fig:experiment-results}. The \gls{uav} performed a satisfactory role exploring the environment and acting as a communications relay node. 

\textbf{Lessons Learned} 
\subsubsection{Physical Layer for Communications}
We note that the communication protocol used in \gls{mocha} does not have any requirement for the physical layer. Nonetheless, we observed experimentally that Rajant radios have a faster association time than ad-hoc WiFi. Rajant radios also provide metrics about the communication links that can be used to trigger communications. This is particularly important for fast \gls{uav} flybys, where slow association time may reduce (or even impede) data transmission between the \gls{uav} and the \glspl{ugv}. 

\subsubsection{Semantics}
The use of semantics provided multiple advantages, from reducing the bandwidth when using our distributed communication architecture to providing invariance between the different viewpoints of the robots. It was a powerful tool for exchanging information between robots with different locomotion modalities. 

\subsubsection{Panoramas}
Our use of depth panoramas as the basis for \gls{ugv} autonomy enabled highly efficient and effective algorithms.
The entire autonomy stack used only around 30\% of the CPU.

\subsubsection{Automation and Operations}
When running multi-robot experiments, the probability of failure compounds with the number of robots. As a consequence, using automated tools for software deployment and updating the robots, such as Docker~\cite{docker}, \texttt{pssh}, and Ansible~\cite{hochstein2017ansible} help reduce mistakes.

\textbf{Opportunities:} 
\setcounter{subsubsection}{0}
\subsubsection{Logistics}
We demonstrated our experiment using a single \gls{uav} and three \glspl{ugv}, but as our simulation experiments indicate, we could benefit from scaling up the number of ground robots. 
Logistics became a central factor here, as our experiment requires a safety pilot per robot. For ground robots, one option is to use \emph{supervisor controllers} that take over the robot when robots deviate from normal operations or try to go into dangerous regions.
\Gls{uav} operations is even more challenging, as pilots should keep line-of-sight with the robot. 

\subsubsection{Extended Operations}
\fer{%
Quadrotors provide stable platforms for aerial mapping. However, these platforms have a limited operation time due to their high energy requirements for hovering~\cite{mulgaonkar2019small}. While \glspl{ugv} may operate for more than two hours, \gls{uav} operation is limited to 30 to 35 minutes under high wind conditions, or when the \gls{uav} velocity is high. Fixed-wing \glspl{uav} may provide avenues for longer mission duration, particularly to act as communication relays. Employing heterogeneous teams of fixed-wing and quadrotors may be an interesting direction for future work.}

\subsubsection{Coordination}
The largest reason for manual takeovers during our experiments was dynamic obstacles, including both vehicles and other robots.
For larger team sizes, these problems would be increasingly severe.
Opportunities therefore exist for not only deconflicting goal selection between \glspl{ugv}, but also planning to avoid potential collisions between robots.

\subsubsection{Traversability}
\begin{figure}
    \centering
    \includegraphics[width=\linewidth]{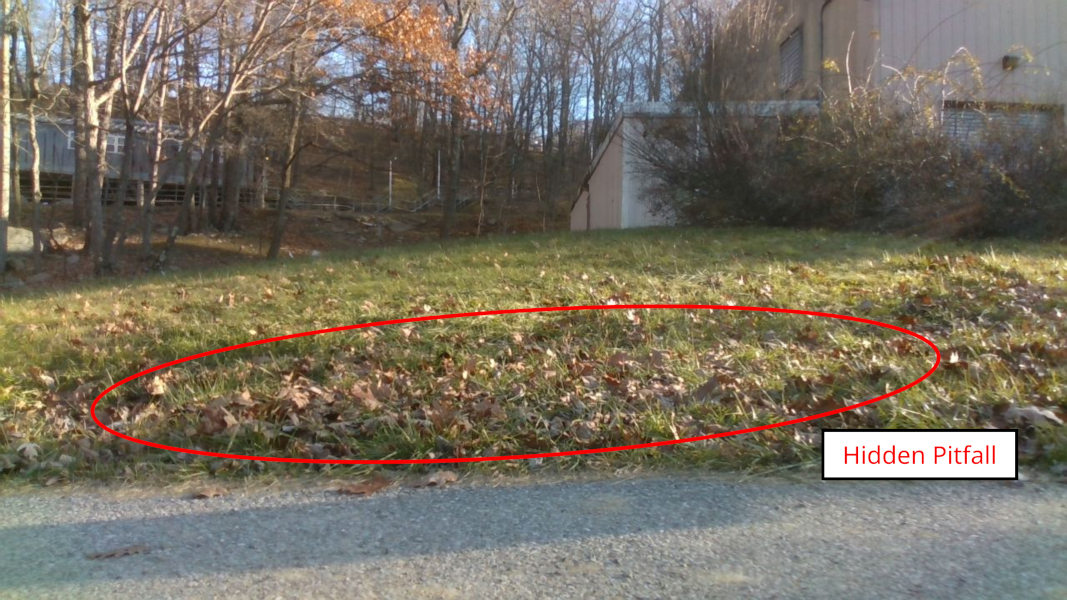}
    \caption{Traversability failure example. The ground robot plans a path across grass, but vegetation and tree leaves hide a pitfall. The \gls{ugv} got stuck, ending its mission.}
    \label{fig:travFailure}
    \vspace{-.4cm}
\end{figure}

In~\cite{miller2024spomp}, we described a method to share traversability information between robots. These traversability methods take into consideration the class in the semantic map, as well as the experience of the robots. However, the local planner and controllers are not modified depending on traversability. One disadvantage of this approach is that robots are prone to exploration of dangerous areas when obstacle free paths are not available, such as the one shown in Fig.~\ref{fig:travFailure}. An improved ground robot controller and local planner could approach uncertain regions more carefully, reducing speeds and planning conservative paths. 
A key challenge here is reasoning about traversability of a certain semantic class in context.
As an example grassy, patches may be traversable most of the time, but at the start of fall season grass and tree leaves may cover hidden obstacles such as ditches.

\section{Open challenges}

Increasing the heterogeneity of our system will further enhance the team's collective sensing, navigation, and communication abilities; these increased abilities will expand the set of operational environments and possible missions. For example, introducing smaller \glspl{uav} such as the Falcon 250~\cite{tao2023seer} would enable rapid sensing in constrained spaces such as dense forests or buildings. 
A larger \gls{ugv} or quadruped would enable navigation through regions difficult for the  Jackals to traverse, such as through tall grass or steep inclines. Effectively increasing the heterogeneity of our system is thus an exciting direction for future work. 
Such research would entail innovations to several key elements of our system, which we discuss below.

\subsection{Traversability Estimation with Heterogeneous Platforms} 
Robots must reliably estimate traversable regions in order to make safe and efficient navigation plans. Traversability estimation in the context of field robotics is especially difficult due to the complex, unstructured, and unpredictable nature of the operational environments. We proposed traversability estimation methods for ground robots in~\cite{miller2024spomp}.
Several recent works propose traversability estimation methods for quadrupeds and wheeled platforms~\cite{cai2023evora, terrainnet, FreyMattamala-RSS-23, jung2023vstrong, shaban2022BEVNet, fan2021step, martinez2020synthetic}.
However, none of these methods provide a mechanism for translating traversability information from one robot to another with different mobility characteristics since most of these works attempt to directly classify the difficulty of crossing the terrain without considering the semantics explicitly.
\cite{borges2022survey} provides an excellent summary of challenges faced in reasoning about traversability from different sensory inputs, and highlight the challenges faced by methods using exteroceptive sensing when a clear geometric signal is absent~(e.g. the possibilty of a ditch under recently fallen leaves).
Intuitively, the information required to assess traversability constraints across varying platforms is highly correlated (e.g., a Jackal and a Husky can both navigate along roads and fields, while only a Husky can navigate steep inclines).
Our previous work incorporates observations made by a team of (homogeneous) \glspl{ugv} to jointly learn a traversabilty estimate over a previously unseen environment~\cite{miller2024spomp}. 
Extending this work to learn platform-specific traversability estimates for a several different platforms by using semantics as a lingua franca to share information would be a key enabler of effective heterogeneous robot teaming, improving efficiency and reliability.


\subsection{Specification and Execution of High-level Tasks} As discussed, we experimentally validate our system in a multi-object search task. 
While this task is representative of a wide range of real-world problems including search and rescue and industrial inspection, robot teams will provide most utility when users can task them with a wide range of missions. 
Current work has focussed largely on indoor environments for convinience of structure and specification in semantically rich settings~\cite{jatavallabhula2023conceptfusion,peng2023openscene,driess2023palm}. It remains unclear the transferability of these methods to km-scale navigation, and the lack of structure in off-road rural navigation settings.
This vision requires a specification framework capable of decomposing a high-level objective into platform specific goals. Ideally, the specification framework will account for the unique abilities and constraints of each platform, such as sensing, traversability, and payload. Recent works considers providing such specifications via formal grammars and natural language~\cite{fang2024high, chen2023scalable}. The proposed methods are only validated in simulation and are executed open-loop. 
Extending this line of research to account for real-time coordination in the presence of noise (in sensing, action, and communication), 
would enable a richer class of missions.

\subsection{Representations for Operations in 3D Environments} 
Shared representations underpin successful planning and coordination of air-ground teaming.
Our system utilize the semantic ortho-map provided by \gls{asoom} as the base representation which which to plan and communicate.
While this representation provides a robust base, it assumes a 2.5D structure that does not easily scale to sharp elevation changes or multi-level structures. 
The perception community has developed many methods for representing complex environments including elevation maps~\cite{chung2024pixel}, object-orientation maps~\cite{sunderhauf2017meaningful}, implicit represenations~\cite{zhi2021place,qiu2024learning} and 3D scene graphs~\cite{erni2023mem, hughes2023foundations, voxgraph, Liu_2023}.
Extending such methods for heterogeneous robot teaming would require scaling to large environments,
fusing observations from different sensing modalities and perspectives, and synchronizing maps across robots.  
Realizing such capabilities would require a significant research effort.
However, as the shared map representation is a limiting factor for heterogeneous system, addressing such challenges would enable operations in a more complex set of environments.

\section{Conclusion}
We have demonstrated effective air-ground teaming using platforms with complementary sensing and mobility constraints.  Our experiments have shown the effectiveness of our system in real-world environments and simulations. The key algorithmic components of our system, namely mapping, planning, and communication, leverage the complementary abilities of each platform for large-scale exploration,
and use semantics as a common framework for communication.

Pursuing the opportunities for improvement and addressing the open research challenges outlined will enable future capabilities of highly heterogeneous air-robot teams for diverse applications.




\bibliographystyle{IEEEtran}
\bibliography{IEEEabrv, literature}

\end{document}